\documentclass[journal]{IEEEtran}
\pdfoutput=1

\makeatletter
\long\def\@makecaption#1#2{\ifx\@captype\@IEEEtablestring%
\footnotesize\begin{center}{\normalfont\footnotesize #1}\\
{\normalfont\footnotesize\scshape #2}\end{center}%
\@IEEEtablecaptionsepspace
\else
\@IEEEfigurecaptionsepspace
\setbox\@tempboxa\hbox{\normalfont\footnotesize {#1.}~~ #2}%
\ifdim \wd\@tempboxa >\hsize%
\setbox\@tempboxa\hbox{\normalfont\footnotesize {#1.}~~ }%
\parbox[t]{\hsize}{\normalfont\footnotesize \noindent\unhbox\@tempboxa#2}%
\else
\hbox to\hsize{\normalfont\footnotesize\hfil\box\@tempboxa\hfil}\fi\fi}
\makeatother

\usepackage{cite}
\usepackage{pbox}
\usepackage{multirow}
\usepackage{footnote}
\usepackage{setspace}
\usepackage[pdftex]{graphicx}
\usepackage{epsfig}
\usepackage{epstopdf}
\usepackage{tabulary}
\newcolumntype{K}[1]{>{\centering\arraybackslash}p{#1}}

\usepackage[cmex10]{amsmath}
\usepackage{mathtools}
\usepackage{mdwmath}
\usepackage{adjustbox}
\usepackage{array}

\usepackage[caption=false,font=footnotesize]{subfig}

\usepackage{fixltx2e}
\usepackage{hyperref}
\begin{document}
\title{The VIP Gallery for Video Processing Education }

%Video and Image Presentation (VIP)
%Video Index Presentation (VIP)
%Video Index Principles (VIP)
%Video in Principle (VIP)
%Video is Processing (VIP)
%Video Instruction Pack (VIP)
%Video and Image Demonstration (VID) Gallery  
%Video and Image Display (VID) Gallery  
%Image and Video Demonstration (Video) Gallery  
%Video Instruction Package (VIP)

%,~\IEEEmembership{Fellow,~IEEE,}
\author{Todd Goodall and
        Alan C. Bovik, Fellow, IEEE
	}

%\doublespacing



%\markboth{IEEE Transactions on Education}{IEEE Transactions on Education}%
%{Shell \MakeLowercase{\textit{et al.}}: Bare Demo of IEEEtran.cls for Journals}




\maketitle

\begin{abstract}

Digital video pervades daily life. Mobile video, digital TV, and digital cinema are now ubiquitous, and as such, the field of Digital Video Processing (DVP) has experienced tremendous growth. Digital video systems also permeate scientific and engineering disciplines including but not limited to astronomy, communications, surveillance, entertainment, video coding, computer vision, and vision research. As a consequence, educational tools for DVP must cater to a large and diverse base of students. Towards enhancing DVP education we have created a carefully constructed gallery of educational tools that is designed to complement a comprehensive corpus of online lectures by providing examples of DVP on real-world content, along with a user-friendly interface that organizes numerous key DVP topics ranging from analog video, to human visual processing, to modern video codecs, etc. This demonstration gallery is currently being used effectively in the graduate class ``Digital Video'' at the University of Texas at Austin. Students receive enhanced access to concepts through both learning theory from highly visual lectures and watching concrete examples from the gallery, which captures the beauty of the underlying principles of modern video processing. To better understand the educational value of these tools, we conducted a pair of questionaire-based surveys to assess student background, expectations, and outcomes. The survey results support the teaching efficacy of this new didactic video toolset. 

\end{abstract}


% \fi
\IEEEpeerreviewmaketitle

\section{Introduction}

\IEEEPARstart{D}{igital} video processing has become ubiquitous given the recent proliferation of mobile devices, significant advancements of video coding standards, fast and practical computer vision algorithms, and breakthroughs in vision science. Finding ways to effectively teach the core fundamental principles of the highly visual and data-intensive field of DVP is quite challenging.

To address these challenges, cross-disciplinary university courses have appeared both in physical and internet classrooms \cite{duke1} \cite{northwest}. Free online courses have provided a practical way for students, faculty, and industry to have access to educational content on their own time. To supplement these online materials, free software resources such as GIMP, FFMPEG, ImageMagick, OpenCV, and Scikit-Image have allowed universally available direct access to common DVP algorithms. However, the availability of code for implementing advanced DVP theories does not translate to a good understanding of DVP principles. Additional tools are needed to provide intuitive, visual, and exciting learning experience enabling complex algorithms to be accessible to inter-disciplinary audiences.

The previously described highly successful Signal, Image and Video Audiovisual demonstration Gallery (SIVA) \cite{rajashekar2002siva} developed in the Laboratory for Image and Video Engineering (LIVE) largely achieved this goal for a collection of image processing topics but only includes three video demonstrations. A user can select any image, select a processing algorithm, refine its parameters, and immediately obtain results. Although SIVA is an excellent resource for demonstrating image processing concepts and techniques, its suite of video processing tools is quite limited, dealing only with a few block motion estimation, frame differencing, and optical flow methods. Owing to hardware processing limitations, memory constraints, and the sheer size of video data, these demonstrations are slow and of limited value.

Given the need for advanced instructional tools for DVP, we have created a new video processing teaching gallery called the Video Instruction Package (VIP), available for download at \cite{VIPdownload}. This gallery provides pre-rendered demonstrations of video processing results divided into 10 categories. Each video includes visual annotations to assist viewers in understanding the theoretical underpinnings of the concepts being taught. An accompanying Graphical User Interface (GUI) interfaces with the VideoLAN Client (VLC) to quickly play any desired demonstration. These hands-on examples will help students to intellectually solidify abstract video-processing concepts.

%--add the number of video demos provided
%--only requires VideoLAN Client (VLC) \cite{VLC} to watch the videos which is free, and the GUI is written in C\# to provide the necessary command line calls to play the RAW video files that have no header information
%--most demo videos provide relevant accomanying information such as descriptional text, functional inputs, mathematical equations, and derivations. Filters are also shown for clarity. This is supplemented by the slides. 

\section{Video Processing Demonstration Gallery}

	In the following sub-sections, the workflow for producing demos is described.  First, a corpus was gathered from high-quality representative real-world video content. Second, low-level video manipulation tools were identified. Third, suitable categories of video processing techniques are described from among those that were selected for classroom demonstration. Further explanation of each category can be found in \cite{completepaper}. 

\subsection{Video Content}

\begin{figure}[!t]
\vspace*{-3.5mm}
\begin{center}
\begin{tabular}{cc}
\subfloat[birdfeeder]{\includegraphics[width=4cm]{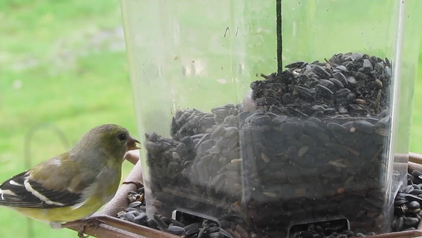}} & \subfloat[cat]{\includegraphics[width=4cm]{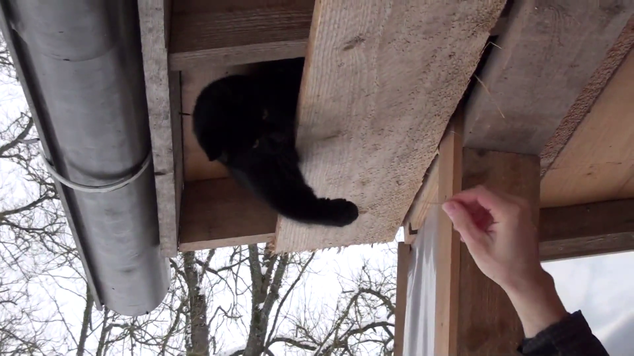}}\\
\subfloat[flower]{\includegraphics[width=4cm]{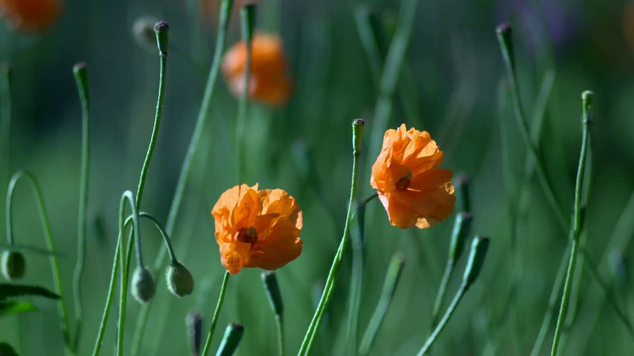}} & \subfloat[football]{\includegraphics[width=4cm]{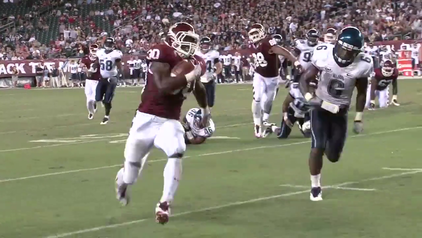}}
\end{tabular}
\end{center}
\vspace*{-2mm}
\caption{Representative frames from the video corpus.}
\vspace*{-4mm}
\label{fig:content}
\end{figure}

	First, a corpus of free and representative video content was gathered. This corpus includes videos having progressive scanning, 1280x720 resolution, and framerates between 24-30 frames per second; all released under the creative commons license. The videos were selected from YouTube and Vimeo, then each video was clipped to a length of 10-15 seconds. Each clip provides a unique scenario in which the camera may shake, pan, or zoom and in which objects may deform, oscillate, or move. Several representative frames from the video collection are depicted in Fig. \ref{fig:content}.  To demonstrate Video Quality Assessment (VQA) models, both pristine and distorted source videos were required. Those videos were obtained from the LIVE Mobile VQA database \cite{DB1}.

\subsection{Development Tools}

To produce visually consistent video content, frames from the video corpus were extracted using the AVCodec library \cite{ffmpeg}, frame rendering was handled by the Simple Direct Media-Layer (SDL) \cite{SDL}, text was rendered using LaTeX, and plots were generated using Python's matplotlib. This workflow allows the use of algorithms written in any language, yielding easier demonstration development.

All videos were rendered into uncompressed YUV 4:2:0 format, to provide high fidelity while avoiding compression artifacts. Playback of this uncompressed video is easily bottlenecked by the physical operation of the hard disk. Thus a H.264 compressed version of each video is also provided in the gallery for ease of playback.

%\begin{figure}[!t]
%\centering
%\includegraphics[width=0.47\textwidth]{images/live_launcher.png}
%\caption{Launcher window in the VIP GUI.}\label{fig:GUI}
%\end{figure}

Finally, a Graphical User Interface (GUI) was designed to organize and present the topics. The GUI layout provides VLC configuration and a main treeview organized by class topics. Each of these topics corresponds to a lecture topic taught in the class ``Digital Video'' offered at The University of Texas at Austin (UT-Austin).

\subsection{Analog Video}

	The UT-Austin class EE 381V ``Digital Video'' emphasizes visual perception, including models of retinal and cortical processing, and this is reflected in the VIP demonstration suite. As an example, Campbell and Robson \cite{spatialtemporalcontrast1} \cite{spatialtemporalcontrast2} studied the contrast responses of the human visual system as a function of spatial and temporal frequency. They traced the spatial and temporal contrast sensitivity functions (CSFs) that are widely used in the design of video codecs and displays. For example, line separation visibility is a function of spatial frequency and stimulus contrast. A demonstration video is provided in which black and white bars (maximum contrast) appear, then slowly shrink to create an increase in apparent spatial frequency. The lines appear to ``merge'' at some point, producing the appearance of an overall constant gray level, demonstrating the upper spatial bandwidth of the human visual system, as predicted by the CSF.

	Video frame rates are generally specified to operate just above the threshold of human vision at which frame flicker becomes detectable. The visibility of flicker is a function of the luminance distribution, spatial frequency profile, and temporal frequency of the stimulus. White frames alternating with black frames are displayed in succession with period length decreasing over time. The video finally cycles black and white at 60Hz, which appears as a gray screen since this flicker frequency falls outside the temporal bandwidth of the vision system, as predicted by the mathematical model of temporal CSF taught in class.

\subsection{Video Singularities and Sampling}

%\begin{figure}[!t]
%\centering
%\includegraphics[width=0.47\textwidth]{images/aliasing_basketball.png}
%\caption{Undersampling and anti-aliasing filters.}\label{fig:aliasing}
%\end{figure}

When sampling analog video in space-time, aliasing artifacts may occur where higher space-time frequencies fold onto lower frequencies. Aliasing is introduced when a signal is undersampled. A video demonstration is provided that shows the result of undersampling then upscaling using the Bilinear, Bicubic, and Lanczos filters.

Digital videos commonly use YCrCb color encoding rather than RGB encoding to exploit the lower frequency response bandwidth to color in the human visual system. Video demonstrations are included for both RGB and YCrCb color spaces.

\subsection{Discrete Video Transforms}

A video demonstration is provided in the VIP gallery that shows losses of information from downsampling and upsampling an input video. Another demonstration video depicts the temporal downsampling and upsampling by multiple factors. Adelson \textit{et al.} \cite{pyramids} designed a multi-scale representation of images called the Gaussian pyramid, widely used in video processing. A video demonstration of this pyramidal decomposition is provided showing 3 scales per video frame.

\begin{figure}[!t]
\centering
\includegraphics[width=0.47\textwidth]{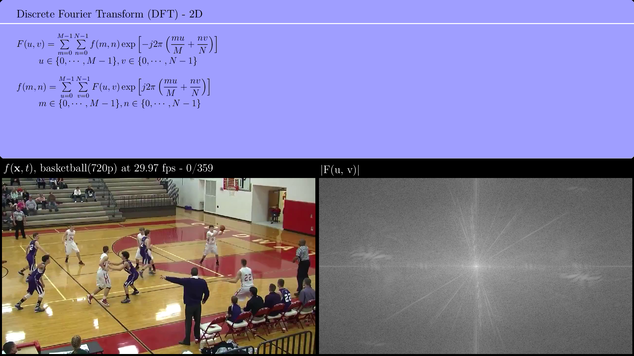}
\vspace*{-2mm}
\caption{2D Spatial DFT.}\label{fig:dft}
\vspace*{-4mm}
\end{figure}

%\begin{figure}[!t]
%\centering
%\includegraphics[width=0.5\textwidth]{images/dct_basketball.png}
%\caption{2D Spatial DCT.}\label{fig:dct}
%\end{figure}

The 2D Discrete Fourier Transform (DFT) and 2D Discrete Cosine Transform (DCT) \cite{DCT} are commonly used for both analysis and decorrelation of the frequency coefficients of digital images. The VIP video demonstrations of 2D (spatial) DFT and centered 2D DCT are depicted in Fig. \ref{fig:dft}. The logarithm of the DFT and DCT magnitudes are displayed to maximize the visibility of the important but subtle structures in the image spectrum. 

%The 2D Discrete Cosine Transform (DCT)  is used to better decorrelate the frequency coefficients of images. While the DFT loses efficiency because it represents spatial discontinuities between periods implied by the frequency sampling, the DCT overcomes this by exploiting a reflection symmetry. A DCT video was developed similar in form to the DFT video depicted in Fig. \ref{fig:dft}. As is done in the DFT demonstration videos, the DCT is also contrast scaled for visibility.

The 2D Discrete Wavelet Transform (DWT) \cite{wavelet} is an efficient decomposition of a frame into pyramidal subbands for analysis and/or perfect reconstruction. A video demonstration of frame DWTs on videos was created that visually exemplifies 3 levels of the DWT pyramid.

\subsection{Video Filters}

%Should I include basic equations?

\begin{figure}[!t]
\centering
\includegraphics[width=0.47\textwidth]{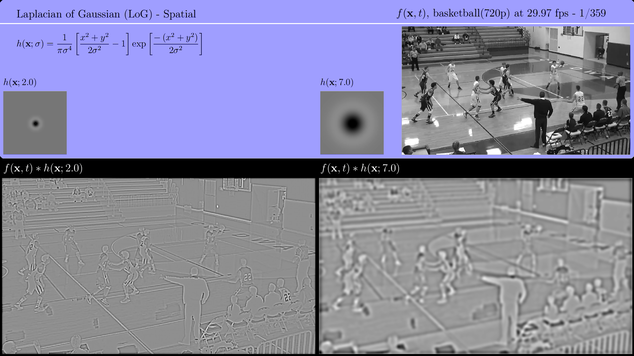}
\vspace*{-2mm}
\caption{Linear spatial frame filtering.}\label{fig:spatial_filter}
\end{figure}

\begin{figure}[!t]
\vspace*{-2mm}
\centering
\includegraphics[width=0.47\textwidth]{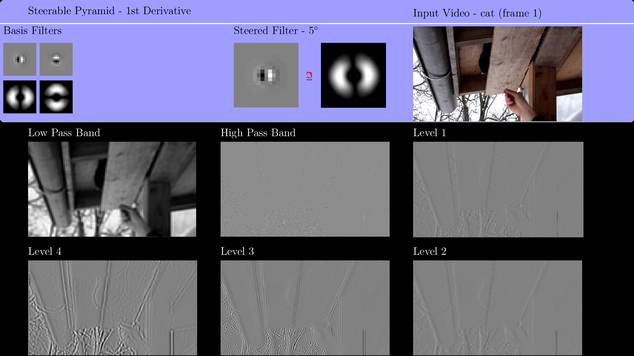}
\vspace*{-2mm}
\caption{Steerable pyramid.}\label{fig:stpyr}
\end{figure}

\begin{figure}[!t]
\vspace*{-2mm}
\centering
\includegraphics[width=0.47\textwidth]{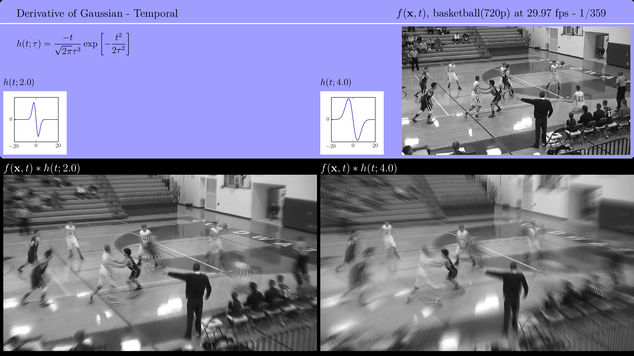}
\vspace*{-2mm}
\caption{Temporal filter example: first derivative of Gaussian.}\label{fig:temporal_filter}
\vspace*{-4mm}
\end{figure}

%also for predictive coding

A number of frame-based linear filters are useful for edge detection \cite{edges}, entropy reduction, and analysis. The 2D Gaussian filter is a low-pass filter that has excellent smoothing properties. The Difference of Gaussians (DOG) filter provides a good model of the responses of retinal neurons. These filters also spatially decorrelate videos and enhance edges. First and second derivatives of Gaussians are directionally bandpass. They provide a scalable, noise-resistant way to compute spatial derivatives. The Laplacian of Gaussian (LOG) shares a similar filter profile to the DOG and is used as an edge detection operator. The ``predictive coding filter'' reduces image entropy, enabling efficient coding of the visual signal. Each of these frame-based linear filters has a video demo, similar to Fig. \ref{fig:spatial_filter}. A 2D visualization of video processing using each filter, with different parameter settings, is included in each demo.

Two groups of filters often used to produce oriented subbands are Gabor filters and steerable pyramid \cite{steerablepyr} filters. Gabor filters produce biologically plausible image decompositions. Each demo video depicts two Gabor filters with different frequency tuning. The steerable pyramid is an overcomplete wavelet decomposition based on derivatives of Gaussians. The video demos, as exemplified by Fig. \ref{fig:stpyr}, show orientated responses over 0-359 degrees, lowpass and highpass residual bands, and 4 scales denoted by Level 1, 2, 3, and 4.

Temporal filters are useful for decorrelation and analysis over time. The VIP suite also has demos of temporal models of perceptually relevant filters including the gamma filter, which is based on a model of the lateral geniculate nucleus (LGN). Video demos are included for temporal first and second derivative of Gaussian filters, Gabor filters, and gamma filters. Fig. \ref{fig:temporal_filter} shows the layout of the temporal processing demos.

\subsection{Motion 1: Detection and Optical Flow}

\begin{figure}[!t]
\centering
\includegraphics[width=0.47\textwidth]{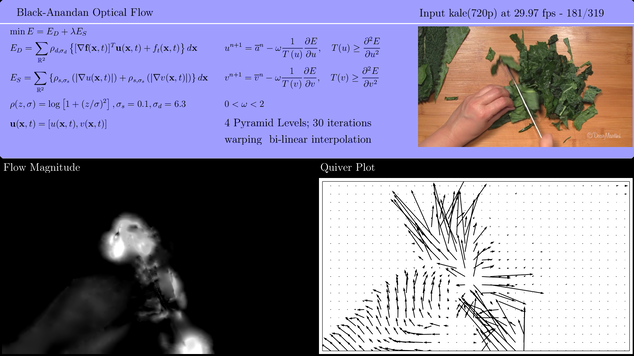}
\vspace*{-2mm}
\caption{Example of Black-Anandan optical flow. Horn-Schunck, Brox \textit{et al.}, and Fleet-Jepson algorithms (not shown) are displayed with similar layout.}\label{fig:flowexample}
\vspace*{-4mm}
\end{figure}

This section of the VIP demo suite focuses on three fundamental optical flow algorithms: Horn-Schunk (HS), Black-Anandan (BA), and Brox \textit{et al.} (BX). The HS algorithm was the first to compute optical flow \cite{horn}. The original HS algorithm assumes that local luminance remains constant between frames and that flow is smooth. The BA algorithm is multiscale, introduces both a non-squared loss function and a median filter to correct erroneous flow vectors \cite{black}. BX solves the optical flow using the calculus of variations \cite{brox} and it uses a gradient constancy assumption. Matlab implementations of HS, BA, and BX developed in \cite{broxcode} \cite{deqing} \cite{deqingcode} were used in the optical flow demos. An example of the layout used for these optical flow algorithms is shown in Fig. \ref{fig:flowexample}. The layout includes annotations reminding the viewer how the flow is computed along with the algorithm-specific parameter settings. 

The gallery also includes a side-by-side comparison of the flow magnitudes from the three methods, letting the user compare flow smoothness and relative accuracy.

\subsection{Motion 2: Perception and Practical Computation}

This section studies models of biological motion perception. A perceptual optical flow algorithm proposed by Fleet and Jepson \cite{fleetj} produces estimates of optical flow using a spatiotemporal filter bank to produce stable estimates of local phase. The base algorithm implemented from \cite{fleetcode} was updated to compute flow on larger images over multiple scales.

\begin{figure}[!t]
\centering
\includegraphics[width=0.47\textwidth]{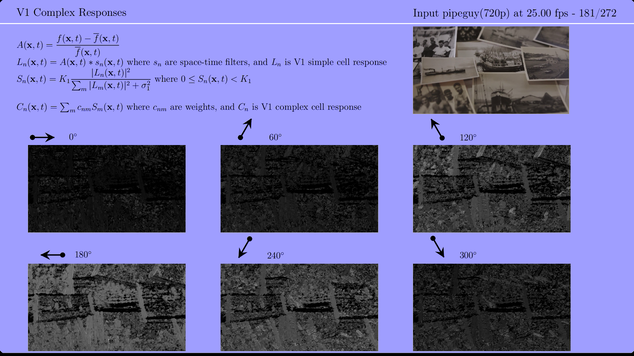}
\vspace*{-2mm}
\caption{Area V1 example.}\label{fig:v1ex}
\end{figure}

As the first stage of human visual information processing, the retina produces a local contrast signal, which is an entropy reducing transformation, used for the models of motion to follow. Annotations are provided with the demo that describe the predictive coding aspect of this contrast signal.

Deeper visual processing of motion is encapsulated in the Simoncelli-Heeger model of brain Area MT \cite{simoncelli1998model} \cite{MTcode}. The presumed neuronal population was set to match the resolution of the HD video inputs. Brain Area V1 neuronal population responses are depicted in Fig. \ref{fig:v1ex} shown with an arrow above each of the 6 directional tunings. To increase the accuracy of the tuned neuronal populations, inhibitory populations were also configured. The processing demo for brain Area MT is configured similarly to the Area V1 demo in regards to layout and orientation selectivity.

\subsection{Statistical Video Models}

\begin{figure}[!t]
\centering
\includegraphics[width=0.47\textwidth]{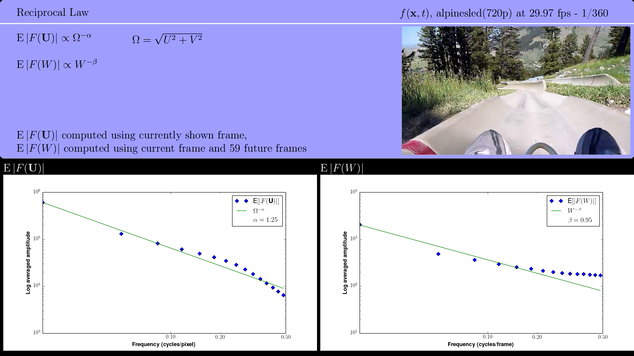}
\vspace*{-2mm}
\caption{Spatial and temporal reciprocal power laws.}\label{fig:rlawalpinesled}
\vspace*{-4mm}
\end{figure}

The next collection of video demos depict and explain regularities in real-world videos that are well-described by ``natural video statistic'' models. The first model illustrated is the reciprocal power law \cite{field1987relations}, which describes the fall-off of the space-time power spectra of videos. As in Fig. \ref{fig:rlawalpinesled}, the demo depicts the relationships between power vs radial spatial frequency and power vs. temporal frequency.

\begin{figure}[!t]
\centering
\includegraphics[width=0.47\textwidth]{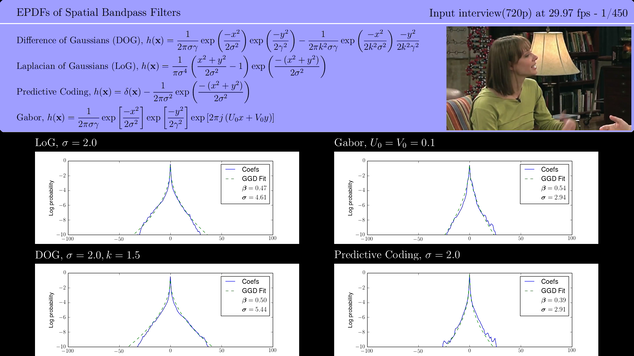}
\vspace*{-2mm}
\caption{EPDFs of spatially filtered frames.}\label{fig:epdf}
\vspace*{-4mm}
\end{figure}

The Empirical Probability Distribution Function (EPDF) of a frame or a video is defined as the normalized histogram. As depicted in the video demos, the EPDF varies widely for different content. However, EPDFs of frames that have been spatially bandpass filtered are highly regular. Indeed, the video becomes nearly decorrelated and gaussianized. As in Fig. \ref{fig:epdf}, this effect is demonstrated with examples of EPDFs of video content filtered by spatial DWT, CFT, Gabor, first and second derivatives of Gaussian, and LOG filters. Similarly, temporal filtering produces this same Gaussian-like regularity, as demonstrated using temporal Gabor, first and second derivatives of Gaussian, and gamma filters.

%\begin{figure}[!t]
%\centering
%\includegraphics[width=0.47\textwidth]{images/lecture7_wf_law.png}
%\caption{Weber-Fechner Law 1.}\label{fig:wflaw}
%\end{figure}

A demo of the Weber-fechner law is also provided. This law describes that a local change in luminance, $\Delta L$, is only visible if $\Delta L>\tau L_{\textrm{ave}}$ where $\tau \approx 0.2$ and $L_{\textrm{ave}}$ is the average local luminance. Using this information, the demo video depicts 9 total patches aligned in a grid with the rows containing 3 discrete average luminance levels and columns containing 3 equally visible amplitudes of flickering dots. The viewer observes that the flickering dots in the left column are easy to detect while the dots in the right column are difficult to detect. The lower left and upper right amplitudes are configured to oscillate at the same luminance, yet the top right dots are not visible due to the relative background luminance. The second demo video includes a constant gradient covered by a sparse grid of dots of oscillating luminance values over the same amplitude range. The viewer should notice vanishing luminance changes toward the right side of the video.

Lastly, spatial contrast masking is demonstrated using sinusoidal gratings overlaid on uniformly distributed random noise, $U \sim [0, 255]$. Two sinusoidal gratings with amplitudes, 20 and 100 are included. As the video plays, the sine grating increases in frequency, with the lower amplitude signal disappearing first as noise visually overpowers the signal.

\subsection{Video Compression}

Video compression standards such as MPEG use the DCT or more advanced transforms such as the H.264 Integer Transform, to decorrelate the video prior to applying lossy compression. The demo video depicts extremes of intra-decoding. The input frame is split into 8x8 sized blocks, transformed to the DCT domain, quantized using MPEG quantization, then inverse transformed to achieve the lossy result. The quantization level is slowly incremented to show how quantization affects the visual quality.

%By contrast to older methods, H.264 uses the integer transform for intra-frame coding. This transformation decouples the transformation from the scaling to produce a more efficient algorithm for the hardware by avoiding computation of irrational numbers by maintaining compression and transformation math in integer domain. Since this produces a similar visual effect as the above, we focus our demonstration on another aspect of H.264, the motion.

%To demonstrate the difference...
%H.264 Scanning and Entropy Coding
%	-show any statistics associeted with this?
%	-modify the quantization and show something similar to above...?

%could easily process the LIVE videos in distorted ways...
%MPEG for LIVE MOBILE (use the same framerate as LIVE below)

The largest differences between the MPEG-1 and MPEG-2 standards are better compression in the latter, support for higher, variable bitrates, and support for interlaced video. H.264 improves on MPEG-2 by providing better compression of HD video content. Pristine videos encoded at multiple bitrates with MPEG-1, MPEG-2, and H.264 codecs are demonstrated to compare visual distinctions.
%Lastly, the VIP gallery includes H.264 compression demonstrations that share the same layout as the MPEG demonstration videos depicted in Fig. 29. Each pristine and H.264 compressed video is obtained from the LIVE Mobile VQA database. The layout allows for comparison among pristine and H.264 compression at the average bitrates 6 Mbps, 2.5 Mbps, and 0.7 Mbps.
%Must mention h.264 and 
%H.264 Video Coding examples
%H.264 is already done for us...
%The LIVE VQA database includes pristine and H.264 compressed versions with varying degrees of quality. A video is made which compares each video in parallel.

\subsection{Video Quality Assessment (VQA)}

\begin{figure}[!t]
\centering
\includegraphics[width=0.47\textwidth]{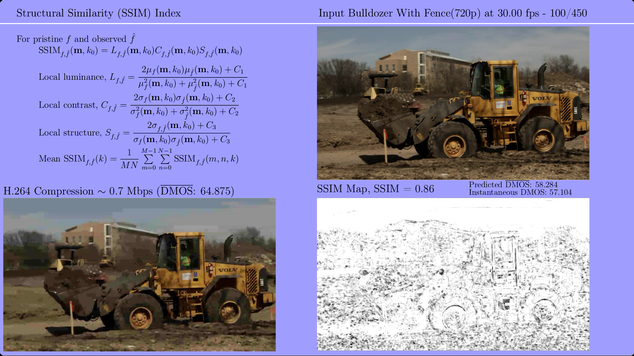}
\vspace*{-2mm}
\caption{SSIM demo comparing pristine and compressed H.264 videos.}\label{fig:ssim}
\end{figure}

\begin{figure}[!t]
\vspace*{-2mm}
\centering
\includegraphics[width=0.47\textwidth]{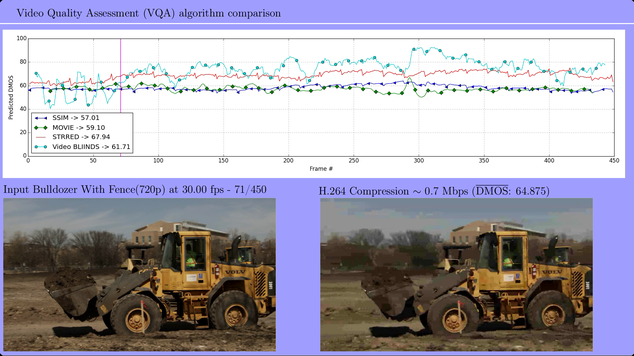}
\vspace*{-2mm}
\caption{Side-by-side comparison of state-of-the-art VQA algorithms.}\label{fig:allqual}
\vspace*{-4mm}
\end{figure}

The demos of state-of-the-art VQA algorithms compare pristine reference videos with distorted versions of them using both Full Reference (FR) and Reduced Reference (RR) algorithms. The Structural Similarity Index Map (SSIM) is an Emmy Award-winning FR model that predicts percieved picture quality by combining luminance, contrast, and structure to produce a local similarity map between distorted and pristine images \cite{ssim}, as shown in Fig. \ref{fig:ssim}. Another top-performing FR metric, the MOtion-based Video Integrity Evaluation (MOVIE) index, measures local motion and spatial accuracy \cite{movie}. Additionally, the MOVIE index produces spatial and temporal quality maps for each frame of input video, both of which are provided in the video demo. The Spatio-temporal Reduced Reference Entropy (STRRED) \cite{soundararajan2013video} index uses reduced reference quality maps to produce quality score predictions. The video demo scales up this reduced map for visualization. Lastly, Video BLIINDS \cite{vqabliinds} is a top-performing No-Reference (NR) algorithm which produces a quality estimate based on feature groups derived from natural video statistic models.

A comparison of state-of-the-art methods is provided as exemplified in Fig. \ref{fig:allqual}. Each frame of input video is scored using SSIM, MOVIE, STRRED, and Video BLIINDS. This allows for study of the strengths and weaknesses of each algorithm over a variety of content.

\begin{table*}[t]
\small
\centering
\caption{Pre-course survey, administered to students within two weeks of the start of the DVP course EE381V. The headings SD, D, N, A, and SA are abbreviations for ``Strongly Disagree,'' ``Disagree,'' ``Neutral,'' ``Agree,'', and ``Strongly Agree.''}
\vspace*{-1mm}
\begin{tabular}{K{0.5in} K{3.8in} K{0.2in} K{0.2in} K{0.2in} K{0.2in} K{0.2in} }
Question & Question Text & SD & D& N& A& SA\\
\hline
\hline
1 & \multicolumn{1}{l}{\scalebox{.9}{I feel that I already have a strong background in fundamental DSP methods.}} & 0 & 3 & 8 & 6 & 0\\
2 & \multicolumn{1}{l}{\scalebox{.9}{I already have a strong background in DIP (have taken a class in it).}} & 1 & 3 & 5 & 7 & 1 \\
3 & \multicolumn{1}{l}{\scalebox{.9}{I have at least some background coursework or exposure to DVP theory.}} & 2 & 6 & 4 & 4 & 1 \\
4 & \multicolumn{1}{l}{\scalebox{.9}{I am familiar with statistical analysis tools like PCA/ICA.}} & 3 & 7 & 4 & 3 & 0 \\
5 & \multicolumn{1}{l}{\scalebox{.9}{I am familiar with models of human visual perception.}} & 2 & 4 & 7 & 3 & 1 \\
6 & \multicolumn{1}{l}{\scalebox{.9}{I am familiar with models of video quality.}} & 2 & 6 & 5 & 4 & 0 \\
7 & \multicolumn{1}{l}{\scalebox{.9}{I like going to movies more than listening to music.}} & 1 & 3 & 7 & 5 & 1 \\
8 & \multicolumn{1}{l}{\scalebox{.9}{I feel that I will learn DV concepts better by seeing them than just doing theory.}} & 0 & 0 & 4 & 4 & 9 \\
9 & \multicolumn{1}{l}{\scalebox{.9}{I am taking this course more as a curriculum need than out of specific interest.}} & 4 & 6 & 4 & 2 & 1 \\
10 & \multicolumn{1}{l}{\scalebox{.9}{I need the ideas from a Digital Video class for my research or outside work.}} & 1 & 2 & 4 & 7 & 3 \\

\hline
\end{tabular}
\label{tab:opinions1}
\end{table*}

\begin{table*}[t]
\small
\centering
\caption{Post-course survey, administered to students within two weeks of the end of the DVP course EE381V. The headings SD, D, N, A, and SA are abbreviations for ``Strongly Disagree,'' ``Disagree,'' ``Neutral,'' ``Agree,'', and ``Strongly Agree.''}
\vspace*{-1mm}
\begin{tabular}{K{0.5in} K{3.8in} K{0.2in} K{0.2in} K{0.2in} K{0.2in} K{0.2in} }
Question & Question Text & SD & D& N& A& SA\\
\hline
\hline
1 & \multicolumn{1}{l}{\scalebox{.9}{I now have a strong background in DVP theory.}} & 0 & 0 & 5 & 11 & 1 \\
2 & \multicolumn{1}{l}{\scalebox{.9}{I am now familiar with models of human visual perception.}} & 0 & 0 & 5 & 8 & 4 \\
3 & \multicolumn{1}{l}{\scalebox{.9}{I am now familiar with models of video quality.}} & 0 & 0 & 2 & 11 & 4 \\
4 & \multicolumn{1}{l}{\scalebox{.9}{I learned DV concepts better by seeing them than just doing theory.}} & 0 & 0 & 1 & 8 & 8 \\
\hline
\end{tabular}
\vspace*{-1mm}
\label{tab:opinions2}
\end{table*}

\subsection{Video Denoising}

Lastly, a demo of a top-performing denoising algorithm, the spatiotemporal Gaussian scale mixture (STGSM) algorithm \cite{stgsm}, provides an all-in-one view of a pristine video, a noisy video, and a denoised output. The equations for the Gaussian scale mixture model are provided for reference in each video.

\section{Questionaire-based analysis of educational effectiveness}

To test the degree to which the VIP video collection could benefit student understanding of the concepts being taught, two surveys based on questionaires were given to the students in the DVP course. The first survey of 10 questions was given at the beginning of the course, and was designed to gauge the students' background preparation and their preconceptions regarding video presentations. The second survey, given at the end of the course, contained 4 questions to determine how much material students learned about Digital Video, as well as how well the video demos facilitated learning.   

The two surveys captured student responses and opinions using a Likert-type scale, meaning that each question could be answered in terms of ``Strongly Disagree,'' ``Disagree,'' ``Neutral,'' ``Agree,'' and ``Strongly Agree.'' The questions that were asked and the results of the first survey are included as Table \ref{tab:opinions1}. From the results, one can conclude that students in this highly cross-disciplinary class come from a wide variety of backgrounds. Surprisingly, although many of the students in the class were Electrical Engineering students, most were unfamiliar with statistical tools such as Principle Components Analysis (PCA) and Independent Components Analysis (ICA). Finally, almost every student agreed that viewing demos of the Digital Video concepts to be taught would assist the learning of Digital Video theory and applications. 

The questions and results of the second survey are included as Table \ref{tab:opinions2}. Notice the shift from the first survey in the level of understanding across topics covered in the class. By comparison to the first survey, more students agreed that the digital video demos aided their understanding, being split evenly between ``Agree'' and ``Strongly Agree.'' The subject with a ``Neutral'' opinion held a ``Neutral'' opinion across all 4 categories for survey 2. Interestingly, this same subject also chose ``Disagree'' for each background question in the first survey, indicating minimal understanding overall.

The student self-assessments are compared between surveys 1 and 2 using a one-tailed Wilcoxon paired signed-rank test for the 16 subjects. Students improved their backgrounds in DVP theory, their backgrounds in human visual perception, and their familiarity with models of video quality, all with high significance ($p < 0.01$). One can conclude that the knowledge gained in the DVP course significantly improved students' relevant background knowledge.

% Total of 16 subjects participated in both surveys.

Using a two-tailed Wilcoxon signed-rank test, no significant difference (p-value of 0.5271) was found for the questions related to seeing DV concepts rather than just doing theory.  For both surveys, subjects selected responses in-between ``Agree'' and ``Strongly Agree.'' Thus, students' expectation that the visual demos helped their learning process was correct.

%% Test statistically if the students had a good background in fundamentals. In survey #1, Questions #1, #2, #3, #4, #5, #6 test background knowledge.

%% Test statistically if the students have a good background in fundamentals after the course. In survey #2, Questions #1, #2, and #3 test this.

%\section{Overview of Course}
%
%The video processing gallery is diverse and covers material mentioned in the EE381V Digital Video Processing graduate course offered at the Univeristy of Texas at Austin. Although it is targetted to be supplemental to the slides which describe the video algorithms in greater detail, the gallery can be greatly useful for anyone working with images and video.

\section{Conclusion}

The LIVE lab envisions that the VIP gallery will be of great use in helping educators adapt their lesson plans for teaching Digital Video in both a more practical and theoretical light. Practical concepts such as applying fundamental algorithms to video will increase the user's intuition.  The opinions gathered from the DVP class show that students prefer to learn by visual example, validating that understanding is strengthened by visual demonstration. The theoretical context provided in most of the gallery videos will also help viewers understand well-known algorithms and should inspire researchers with ideas of how to improve these fundamental techniques.

\section*{Acknowledgment}

The students of EE381V are acknowledged for providing incredible feedback regarding the demonstration content.

\ifCLASSOPTIONcaptionsoff
  \newpage
\fi



\bibliographystyle{IEEEtran}
%\begin{thebibliography}{1}
\bibliography{main}

\end{document}